\documentclass{article}

\usepackage{arxiv}
\usepackage{threeparttable}
\usepackage{multirow}
\usepackage[utf8]{inputenc} % allow utf-8 input
\usepackage[T1]{fontenc}    % use 8-bit T1 fonts
\usepackage{amsmath}
\usepackage{hyperref}       % hyperlinks
\usepackage{url}            % simple URL typesetting
\usepackage{booktabs}       % professional-quality tables
\usepackage{amsfonts}       % blackboard math symbols
\usepackage{nicefrac}       % compact symbols for 1/2, etc.
\usepackage{microtype}      % microtypography
\usepackage{cleveref}       % smart cross-referencing
\usepackage{lipsum}         % Can be removed after putting your text content
\usepackage{graphicx}
\usepackage{natbib}
\usepackage{doi}
\usepackage{algpseudocode}
\usepackage{algorithm}

\title{UI Layers Merger: Merging UI layers via Visual Learning and Boundary Prior}

\author{ 
    {Yun-nong Chen}\\
	School of Software Technology\\
	Zhejiang University\\
	\And
	{Yan-kun Zhen}\\
	Alibaba Group\\
	Hangzhou, China
	\And
	{Chu-ning Shi}\\
	College of Computer Science and Technology\\
	Zhejiang University\\
	\And
	{Jia-zhi Li}\\
	College of Computer Science and Technology\\
	Zhejiang University\\
	\And
	\href{http://orcid.org/0000-0002-9049-0394}{\includegraphics[scale=0.06]{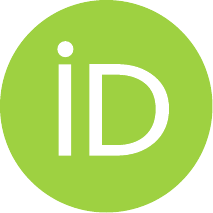}\hspace{1mm}Liu-qing Chen} \\
	College of Computer Science and Technology\\
	Zhejiang University\\
	\And
	{Ze-jian Li}\\
	School of Software Technology\\
	Zhejiang University\\
	\And
	{Ling-yun Sun}\\
	College of Computer Science and Technology\\
	Zhejiang University\\
	\And
	{Ting-ting Zhou}
	Alibaba Group\\
	Hangzhou, China\\
	\And
	{Yan-fang Chang}
	Alibaba Group\\
	Hangzhou, China\\
}

\renewcommand{\shorttitle}{\textit{arXiv} Template}

\hypersetup{
pdftitle={UI Layers Merger: Merging UI layers via Visual Learning and Boundary Prior},
pdfsubject={q-bio.NC, q-bio.QM},
pdfauthor={Yun-nong Chen, Liu-qing Chen},
pdfkeywords={UI Layers Merger: Merging UI layers via Visual Learning and Boundary Prior},
}

\begin{document}
\maketitle

\begin{abstract}
	With the fast-growing GUI development workload in the Internet industry, some work attempted to generate maintainable front-end code from UI screenshots. It can be more suitable for utilizing UI design drafts that contain UI metadata. However, fragmented layers inevitably appear in the UI design drafts which greatly reduces the quality of generated code. None of the existing GUI automated techniques detects and merges the fragmented layers to improve the accessibility of generated code. In this paper, we propose UI Layers Merger (UILM), a vision-based method, which can automatically detect and merge fragmented layers into UI components. Our UILM contains Merging Area Detector (MAD) and a layers merging algorithm. MAD incorporates the boundary prior knowledge to accurately detect the boundaries of UI components. Then, the layers merging algorithm can search out the associated layers within the components' boundaries and merge them into a whole part. We present a dynamic data augmentation approach to boost the performance of MAD. We also construct a large-scale UI dataset for training the MAD and testing the performance of UILM. The experiment shows that the proposed method outperforms the best baseline regarding merging area detection and achieves a decent accuracy regarding layers merging.
\end{abstract}

% keywords can be removed

\section{Introduction}
Graphic User Interface (GUI) is an important visual communication part that can play an essential role in the app's success. UI design draft is a high-fidelity prototype of GUI, and it has a view hierarchy representing both in how UI components are constructed and how they are arranged. One of the main jobs of a front-end engineer is to implement the code. With today's increasing development in Internet industry, there is a huge demand for front-end code development. To relieve front-end developers from tedious and repetitive work, some previous research has adopted automatic methods to generate maintainable code from UI screenshots\citep{beltramelli2018pix2code, behrang2018guifetch}. But it can be more suitable to generate code from UI design drafts that contain the UI metadata. The Imgcook\footnote{\url{https://www.imgcook.com/}} is such a tools that can automatically generate front-end code from UI design drafts.

To generate high-quality and maintainable front-end code with automatic code generation tools, the UI design drafts need a concise and structured view hierarchy. In practice, designers usually produce a UI design draft by facilitating the overlay of layers which represent basic shapes and visual elements, with design software such as Sketch\footnote{\url{https://www.sketch.com/}} and Figma\footnote{\url{https://www.figma.com/}}. These fragmented layers without structured grouping inevitably increase the difficulty of understanding the semantics of UI components and also impairs the maintainability of the generated code. 
%fig 1
\begin{figure}
\centering 
\includegraphics[scale=0.10]{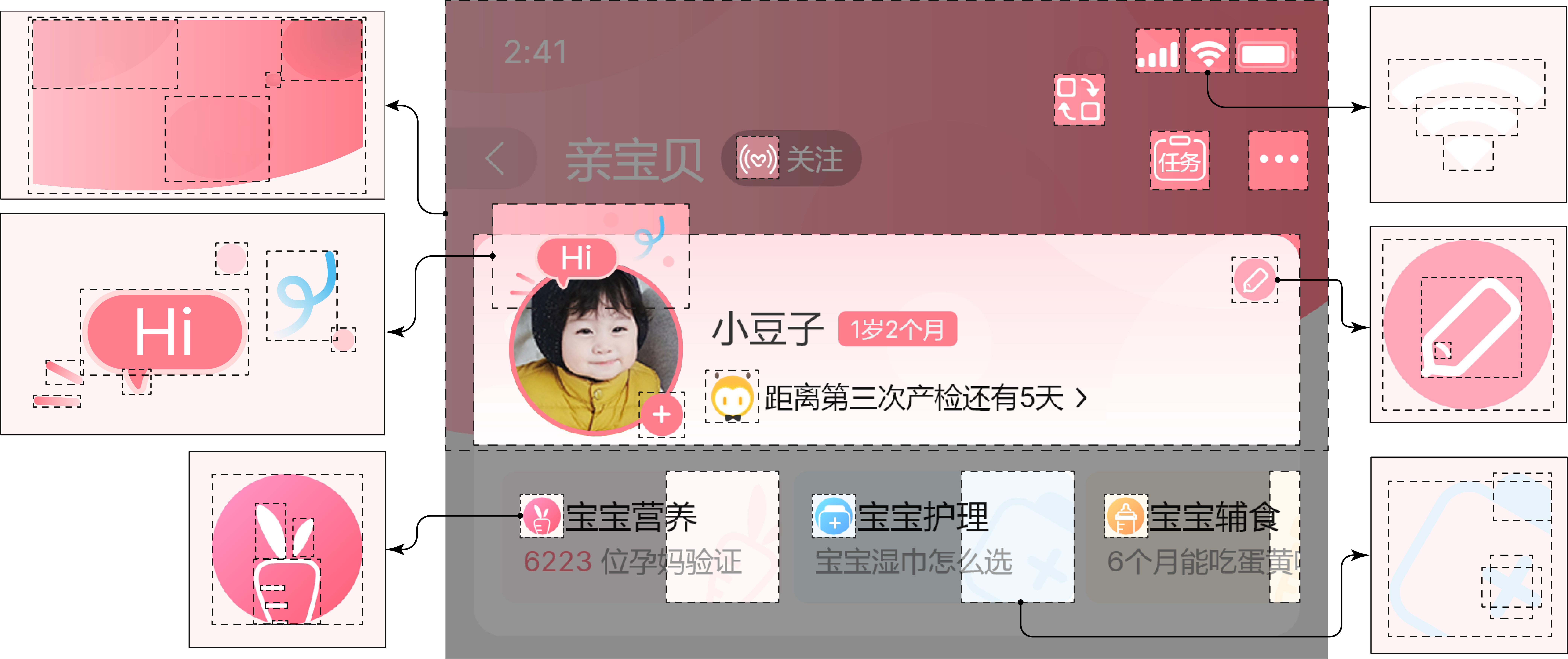}
\caption{Examples of fragmented layers in UI components. The dashed rectangles represent the fragmented layers. These fragmented layers form UI components with semantic information.}
\label{fig1}
\end{figure}

\begin{figure}
\centering 
\includegraphics[scale=0.55]{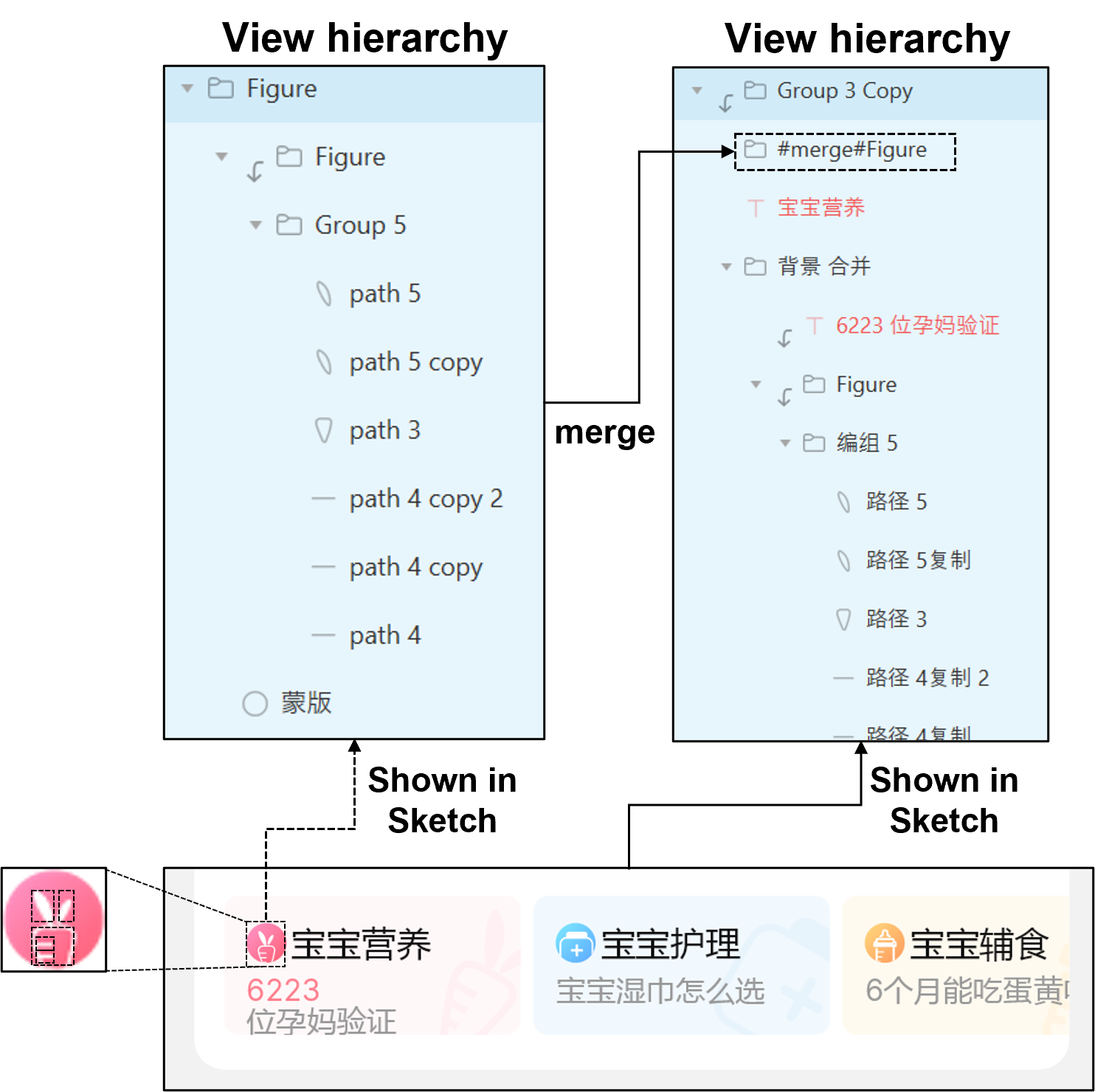}
\caption{An example of merging fragmented layers. The dashed boxes in the "carrot-like" icon represent fragmented layers which are shown in the view hierarchy on the top-left side under the "Figure" folder. After merging these fragmented layers, the view hierarchy on the top-right side is simplified as one single "\#merge\#" container.}
\label{fig:expand_issue}
\end{figure}
% \begin{figure}[!htb]\small
% \centering
% \begin{tabular}{c}
% \includegraphics[scale=0.1]{img/fig_issue.png}\\
% {\footnotesize\sf (a) Examples of fragmented layers in UI components} \\[3mm]
% \includegraphics[scale=0.55]{img/fig_expand_issue.png}\\
% {\footnotesize\sf (b) Examples of merging fragmented layers} \\
% \end{tabular}
% \caption{Examples of UI fragmented layers. }
% \label{fig1}
% \end{figure}

In a code generation process, the associated fragmented layers need to be merged into UI components to avoid increasing the complexity of the hierarchical structure and even reduce the quality of generated code. As shown in Fig.~\ref{fig1}, the dashed rectangular boxes represent the fragmented layers. It can be noticed that some fragmented layers, together to represent one UI component. As shown in Fig.~\ref{fig:expand_issue}, the view hierarchy without merging fragmented layers in the UI icon is complicated and redundant. After the merging, the layout structure is simplified greatly as we can use a single container to represent the component. However, due to the substantial number of layers and complex design patterns in a UI design draft, it is time-consuming for designers or developers to locate and merge multiple fragmented layers manually. In this study, we try to locate all the UI components, of which the associated fragmented layers need to be merged. Then we can merge these layers into UI components. Therefore, the quality of generated code can be improved as the complexity of the view hierarchy is greatly reduced.

As described above, the challenge is to determine the location and amount of UI components in a design draft. Given an arbitrary UI design draft, the amount of UI components is various. The UI components have various types, such as icons, atmosphere UI, and background UI. Multiple types result in different sizes and aspect ratios. For example, there is a significant size difference between the background UI (top-left) and the WiFi icon (top-right) as shown in Fig.~\ref{fig1}. Another challenge is that we need to search out all the associated fragmented layers in the region of UI components when their locations are found. Fig.~\ref{fig1} shows the atmosphere UI (middle-left) consists of six fragmented layers and we have to find all of them accurately. Intuitively, the more accurate the detected boundaries of UI components are, the more accurately we can find the associated layers.

In particular, our method solves the problem in two steps: merging area detection and layers merging, which are implemented by the Merging Area Detector (MAD) and the layers merging algorithm respectively. MAD can automatically detect the areas of UI components. The layers merging algorithm uses the located merging area to merge the fragmented layers into UI components. Before feeding data into MAD, we construct a preprocessing pipeline that parses the UI design drafts to obtain the screenshots with view hierarchy. The boundary information of layers in view hierarchy can be incorporate to MAD as prior knowledge. Additionally, a novel data augmentation and spatial fusion strategy are introduced to boost the performance of our MAD. To train our MAD and evaluate the effectiveness of our UILM, we collect and construct a UI dataset with modern diverse UI design drafts.

% approach
\begin{figure*}
\centering 
\includegraphics[scale=0.5]{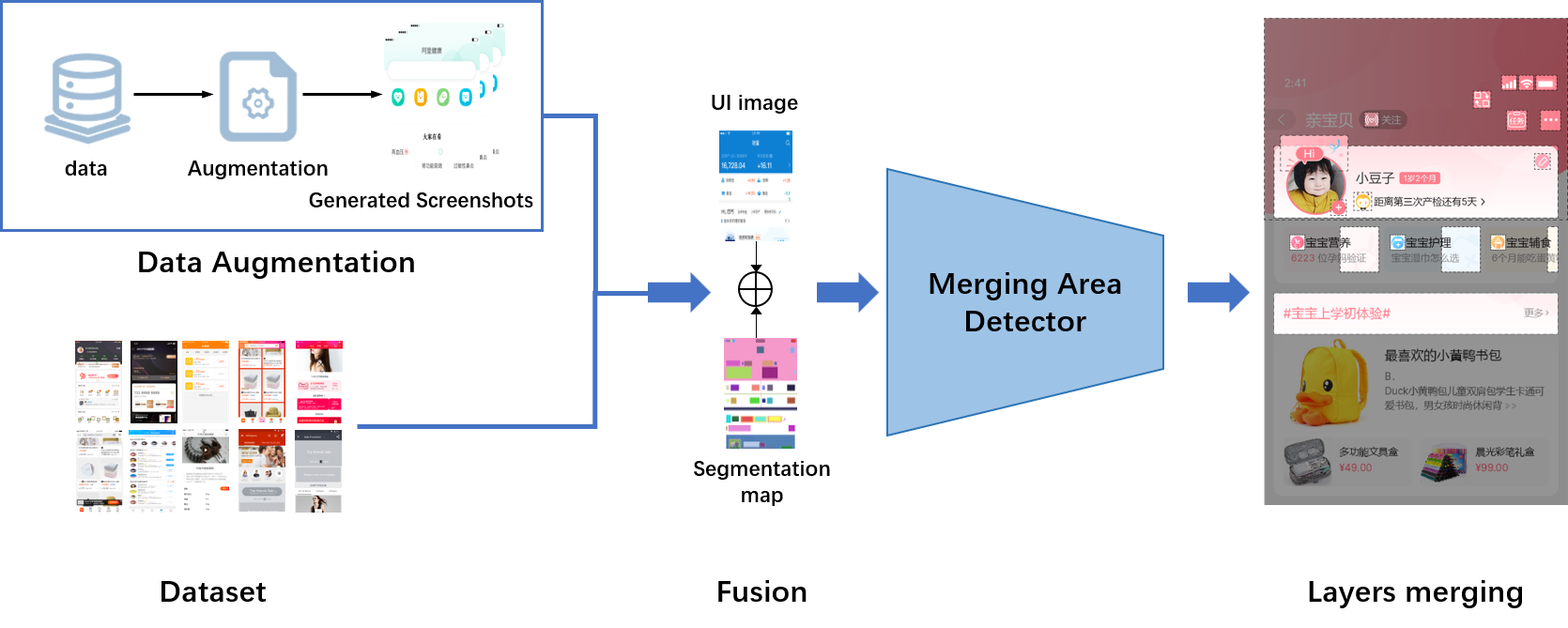}
\caption{The overview of proposed method.} \label{fig2}
\end{figure*}

We summarize the contributions of this study as follows:
\begin{enumerate}
\itemsep -1pt
\item We construct a large UI dataset consisting of UI design drafts. A dynamic data augmentation approach and spatial fusion strategy is introduced to boost the performance of our method.
\item This is the first work to solve the fragmented layers issue by proposing a method called UILM, which contains Merging Area Detector and a layers merging algorithm.
\item Our MAD outperforms the best baseline regarding merging area detection. The experiments results based on three specific development conditions show that our method can successfully support automatic code generation. The user study on real application also confirms the effectiveness of our UILM.
\end{enumerate}

\section{Related Work}
\label{sec:headings}

\subsection{UI Code Generation}
To free the developer from tedious and repetitive work, recent research on automatically generating code from various design stages has been conducted, including hand-drawn sketches \citep{SuleriPSJ19, acsirouglu2019automatic, jain2019sketch2code}, wireframes \citep{Ge19,  halbe2015novel}, and GUI screenshots \citep{beltramelli2018pix2code, chen2018ui, nguyen2015reverse, moran2018machine, FengMYCZZ21}.  \cite{SuleriPSJ19} propose the approach for web code generation process from hand-drawn mock-ups using computer vision techniques and deep learning methods. Sketch2Code \citep{jain2019sketch2code} employs DNN to detect GUI elements in sketches. Its output is a platform-independent UI representation object used by a GUI parser to create code for different platforms. However, the generation from sketches or wireframes to code still requires to manually modify the generated code. To improve the usability of the generated code, Pix2code \citep{beltramelli2018pix2code}, which is based on CNN and RNN, can generate code from a GUI screenshot. To generate more appropriately structured code, \cite{chen2018ui} present a neural machine translator that combines computer vision and machine translation techniques for converting a UI design image to a GUI skeleton.

The above mentioned UI code generation techniques do not fully utilize UI metadata, such as view hierarchies or accessibility tags, thus some work improve the quality of generated code by taking the UI design drafts to generate more useful and maintainable front-end code. Yotako\footnote{\url{https://yotako.io/}} and Imgcook are such tools that take UI design drafts as input which are made by designers with modern design software. The resulting code from these tools describes the original UI using only relative constraints, allowing it to be responsive to different front-end platforms, e.g., Vue, React, Angular, etc. However, the existence of fragmented layers misleads the machine so that there are many fragmented containers in the generated code. The generated code often does not satisfy developers' requirements due to the lack of maintainability. We seek to address this limitation and advance the body of work on code generation with our approach.

\subsection{UI Detection and Dataset}

The first step of our method is to use UI screenshots and raw view hierarchies to detect the regions of the fragmented layers. Here we briefly survey existing GUI elements or components detecting techniques. Some work \citep{DBLP:conf/sigsoft/ChenXXCXZ020, chen2019automated, moran2018machine} first use traditional image processing methods, such as edge detection, to locate UI elements, and then applies the CNN model to identify the semantics of UI elements (e.g., UI type). \cite{LiuCWHHW20} present the Owl Eyes to classify and detect the GUI display issues, such as missing images and text overlap, for guiding the developers to fix these bug. They further develop a fully automated approach, Nighthawk \citep{liu2022nighthawk} which is based on the Faster RCNN model, to detect GUIs with display issues and locate the detailed region of the issue for guiding developers to fix bug. GalleryDC \citep{chen2019gallery} uses the Faster RCNN to detect UI components and automatically create a gallery of GUI design components. \cite{white2019improving} utilize YOLOv2 to identify GUI widgets in UI screenshots to improve GUI testing. \cite{ZhangGSWMYSNWFE21} collected and annotated a large UI dataset from iPhone apps to train a robust and efficient on-device model to detect UI elements. 

To train a model for detecting and merging fragmented layers in UI design drafts, a large-scale dataset consists of UI design drafts is mandatory and important. More recently, Rico \citep{Rico}, a large-scale data of Android apps, is generally used as the data source of related research on UI. Different from the traditional image-based dataset, Rico consists of about 72,000 UI examples from 9722 Android apps. Each example is associated with a screenshot of particular UI design, the corresponding view hierarchy, and the user interaction information. It marks 24 UI component types, 197 text button concepts, and 97 icon types on these view hierarchies. However, it has some limitations, such as broken hierarchies, inaccurate bounding box boundaries and  inconsistencies with the class label  of similar objects \citep{VINS}. Although \cite{LearningtoDenoise} develop tools for improving RICO dataset quality, more complex UI layouts and UI design specifications have been a challenge for UI research. \cite{ZhangGSWMYSNWFE21} attempt to fill this gap by collecting and annotating a large UI dataset from modern iPhone apps. Hence, it is important to have a well-structured dataset that can provide more accurate and more diverse UI design to support UI research. Besides, to the best of our knowledge, there is no public dataset for the task of detecting fragmented layers. Therefore, we collected UI drafts and constructed a UI dataset with modern diverse UI components that enables to train a fragmented layers merging area detector.

In this paper, we propose a novel object-detection based method to detect the regions of the fragmented layers in UI design drafts. Given that UI design drafts has view hierarchies, we enrich visual features by incorporating the layers' boundary information.

\section{Methodology}
\label{sec:others}
In this section, we introduce the details about the proposed method. As illustrated in Fig.~\ref{fig2}, we present a dynamic data augmentation approach, and design the Merging Area Detector (MAD). The segmentation map encodes the boundary information and the location of the UI components. By adding the segmentation map to the UI images, MAD can utilize spatial information of layers to condition the proposal bounding features. The layers merging algorithm can merge fragmented layers into UI components. After merging the layers, they can be transformed into one single image type layer which can significantly reduce the complexity of layers in the UI design drafts and improve the quality of generated code.

%fig 3
\begin{figure*}
\centering 
\includegraphics[scale=0.7]{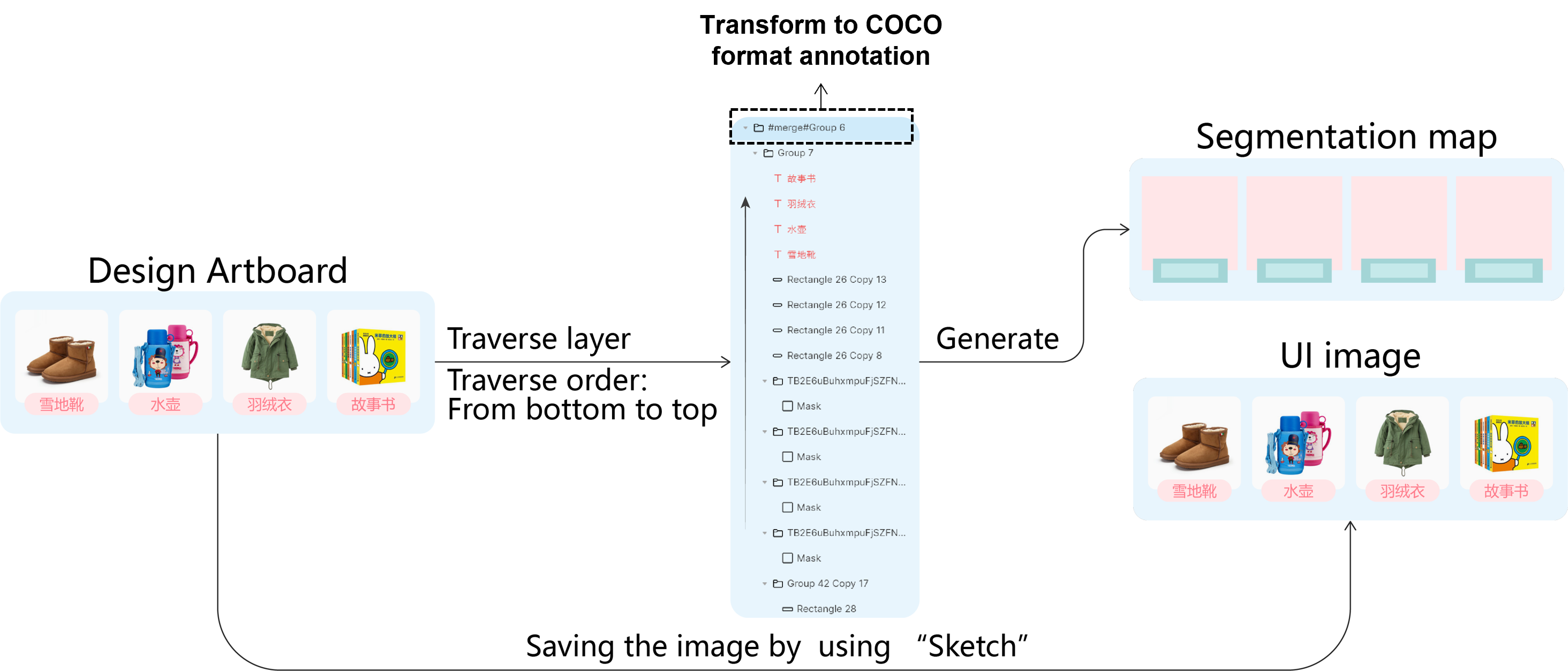}
\caption{The parsing pipeline of UI drafts preprocessing.} 
\label{fig3} 
\end{figure*}

\subsection{Dataset Construction}

To investigate our automatic approach for merging fragmented layers in design drafts, it is essential to have a large-scale, carefully annotated dataset of UI design drafts. Thus, we create the UILM dataset\footnote{\url{https://github.com/zju-d3/UILM}}, a novel well-annotated dataset of over 600 Sketch files. Each Sketch file consists of several design drafts. Each design draft has a view hierarchy to organize layers. A view hierarchy is a tree structure where each layer in the tree corresponds to an element in the UI. Different from RICO \citep{Rico}, our single layer represents the minimum element of an arbitrary UI object. In fact, in design drafts, it is very common for multiple fragmented layers to form a UI component, such as icons and background UI.

When constructing the dataset, we recruited UI designers to perform data annotation. They were asked to read a document that detailed the steps for data annotation. Then they were provided an example set of annotated UI design drafts where all the fragmented layers are clustered into several groups and associated with the "\#merge\#" label. This provided a better understanding of how to make a right "\#merge\#" annotation. Eventually, they used Sketch to manually locate fragmented layers at given UI design drafts and manually merge these layers into groups. In consequence, these groups were named starting with "\#merge\#".

In the data preprocessing stage, given an annotated UI design draft, we parsed its artboard to a JSON file and generated the train labels with COCO format \citep{lin2014microsoft}. The parsing pipeline is shown in Fig.~\ref{fig3}. The Sketch tools saved the corresponding screenshots. The layers in an artboard could be regarded as a view tree. We traversed the tree from bottom to top and obtain the spatial information of all layers. The information of each layer included position, size, type, fill, etc. When we traversed to the layer group containing “\#merge\#” label, we made the training data $\left [ x,y,w,h \right ]$ based on the previously obtained position and size information. To highlight the boundary information, we generated segmentation maps by filling the layers with colors in the traversing order. In summary, with UI drafts, we have generated UI screenshots, corresponding segmentation maps, and training data containing groups of "\#merge\#" labels and layers' information.

\subsection{Merge Area Detection}

\begin{figure*}
\centering 
\includegraphics[scale=0.66]{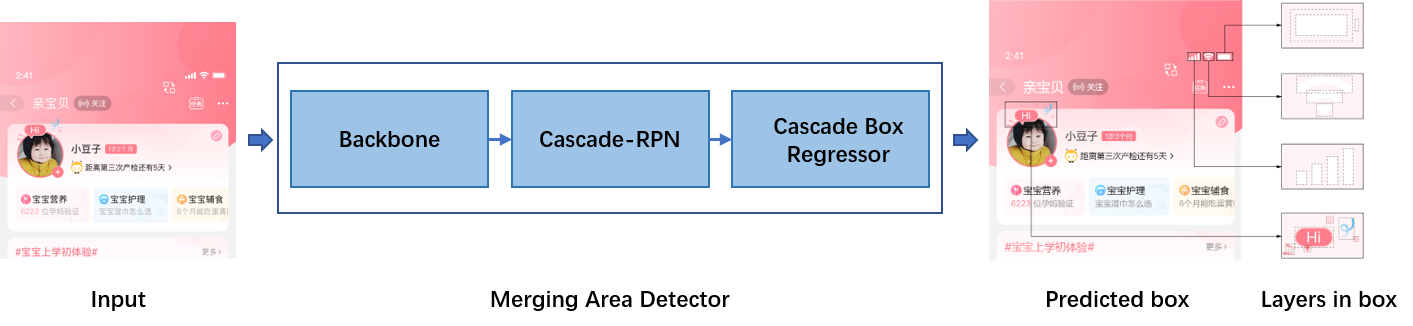}
\caption{The architecture of Merging Area Detector.} 
\label{fig4} 
\end{figure*}

We need to locate the merging area of the UI component and determine reliable boundary for the area. It's difficult to merge layers with high accuracy if the predicted boundary is not accurate enough. As illustrated in Fig.~\ref{fig4}, to solve this issue, we adopt multi-stage adaptive convolution\citep{vu2019cascade} to learn an anchor with adaptive shape to improve the boundaries alignment accuracy in the context of complex UI layout.

%fig 4 Detector Architecture

Specifically, given a feature map $x$, we calculate each location $p$ on the output feature $y$ in the adaptive convolution as follows: 
\begin{equation}
    y\left [ p \right ] =\sum_{ O\in \Omega } \omega\left [ O\right ] \cdot x\left [ p+O \right ] 
\end{equation}

where $\Omega$ is the offset field and offset $O$ can be decomposed into center offset and shape offset, which can be obtained by:
\begin{equation}
    O=O_{ctr}+O_{shp} 
\end{equation}

where $ O_{ctr}=\left ( \bar{a}_{x} - p_{x} , \bar{a}_{y} - p_{y} \right ) $. $O_{shp}$ is defined by anchor shape and kernel size. $\bar{a}$ denotes the projection of anchor $a$ onto the feature map.

At region proposal network, to select sufficient positive samples and avoid to establish a loose requirement for positive samples, we use two stages by starting out with an anchor-free metric followed by anchor-based metric in the second stage. In the first stage, the adaptive convolution is set to perform dilated convolution since anchor center offsets are zeros. Its function is to increase the feature map perception field. In the second stage, we compute the anchor offset $O^{\tau}$ and fed it into the regressor $f^{\tau}$ to produce the regression prediction $\hat{\delta }^{\tau}$. The prediction $ \hat{\delta }^{\tau} $ is used to produce regressed anchors $ a^{\tau+1} $. The objectness scores are then produced from the classifier, followed by NMS to produce the region proposals.

At box regressor, improving the layers boundary alignment rate depends on box regression performance of the detector. In Faster-RCNN \citep{ren2015faster}, a fixed threshold IoU is set, typically $u=0.5$, which establishes quite a loose requirement for positives. This makes it difficult to train a detector to achieve optimal performance. If the IoU is set higher, typically $u=0.7$, which leads to a few positive samples after filtering and causes the model to be over-fitting. This paper follows the suggestions from \citep{cai2018cascade} that a detector optimized at a single IoU level is not necessarily optimal at other levels. In other words, each IoU level can learn an optimal bounding box regressor independently. We adopt the cascade box regressor that improves performance by multi-stage expansion while avoiding the over-fitting problem.

%%% 具体的合并规则算法
\subsection{Layers Merging Algorithm}

Our ultimate goal is to find redundant layers from the layers list and merge them to UI components. So after detecting the merging areas from screenshots, we need to trace back to the corresponding layers from UI drafts in the detected merging area. We propose the layers merging algorithm that merge the relevant layers while filtering the irrelevant layers in the detected area.

The complete merging process is shown in Algorithm~\ref{algorithm_merge}. With the JSON file and the predicted bounding box list as the input, we first flatten the hierarchical layers to a layers list, since it is assumed that the tree hierarchy from UI designers is not reliable. We then index the layers list and arrange the predicted bounding boxes in ascending order. 

Given a predicted bounding box, we traverse the layers from the flatten list $fl$. The area of layers can be calculated. We calculate the intersection area of the given predicted bounding box and all layers. The layers that exceed the pre-determined area threshold $T_i$ are be saved to the filtered list $fi$. We index the layers by their absolute position in the list. We can calculate the distance between any two layers by using the index. The mean distance of filtered layers is used as the distance threshold $T_d$. If the distance between adjacent layers is below the threshold $T_d$, we save the layer to result group $res$. Finally, the saved layers are removed from the flatten layers list, and the flatten layers list is updated. After processing all predicted bounding boxes, we find out all the fragmented layers belonging to particular UI components.
% 合并算法
\begin{algorithm}\small
\centering
\caption{Layers merging} 
\label{algorithm_merge}
\begin{algorithmic}[1]
    \Require {$N$ predicted bounding boxes ${\left \{ bb_{i} \right \} }_{i=1}^{N} $ and their JSON files.}
    \Ensure {result group $res$.}
    \State{$T_i \gets \text{pre-determined\ threshold\ of\ the\ intersection}$;}
    \State{Traverse JSON files to obtain the flatten layers list $\left \{ fl_{j}  \right \} _{j=1}^{M}$ and obtain the index of the layers;}
    \State {Arrange ${\left \{ bb_{i} \right \} }_{i=1}^{N}$ in ascending order;}
    \For {\textbf{all} $bb_i$ in $bb$}
        \For {\textbf{all} $fl_{j}$ in $fl$}
            \If {$fl_{j} \cap bb_i > T_i$}
                \State {save the layer $fl_{j}$ to the filtered list $\left \{ fi_{k}  \right \} _{k=1}^{K}$;}
            \EndIf
        \EndFor
        \State {Compute the distance threshold $T_d$;}
        \For {\textbf{all} $fi_{k}$ in $fi$}
            \If {${fi_{k+1}.index - fi_{k}.index} < T_d$}
                \State {save the layer to result group $res$;}
            \EndIf
        \EndFor
        \State {remove the layers in $res$ from flatten list $fl$ and update $fl$;}
    \EndFor
    \Return {$res$.}
\end{algorithmic}
\end{algorithm}

\subsection{Feature Fusion}

Most of the UI design drafts in our research are from mobile online shopping platforms, which have pretty rich semantic information, such as components, icons, and backgrounds. Besides, a variety of UI components in e-commerce scenarios leads to complex boundary information in constructed layers, which makes it difficult to learn the bounding spatial features. So we propose a fusion strategy to utilize the spatial information as prior knowledge. 

\begin{figure*}[!htb]\small
\centering
\begin{tabular}{c}
\includegraphics[scale=0.6]{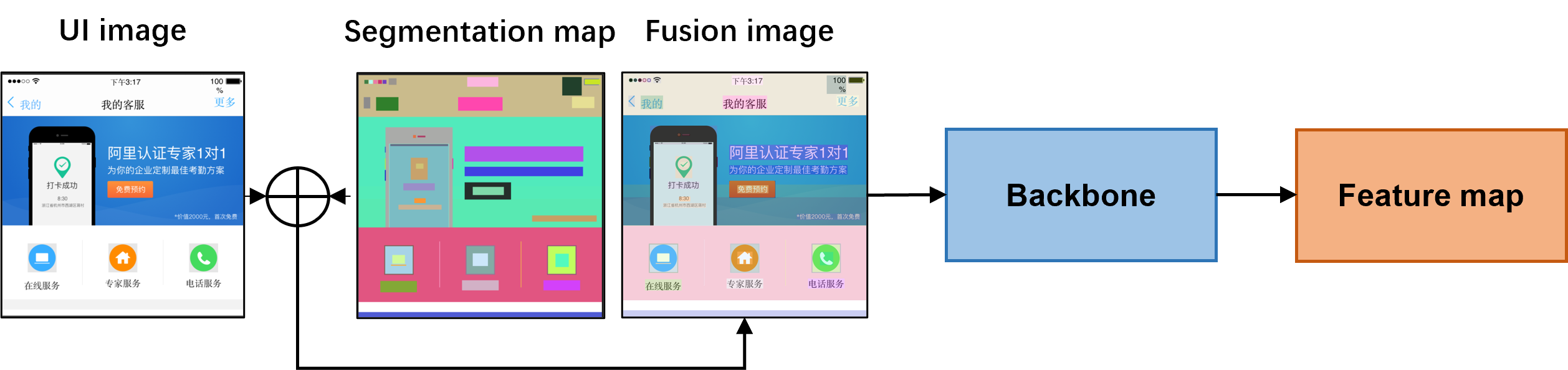}\\
{\footnotesize\sf (a) The spatial fusion strategy.} \\
\includegraphics[scale=0.6]{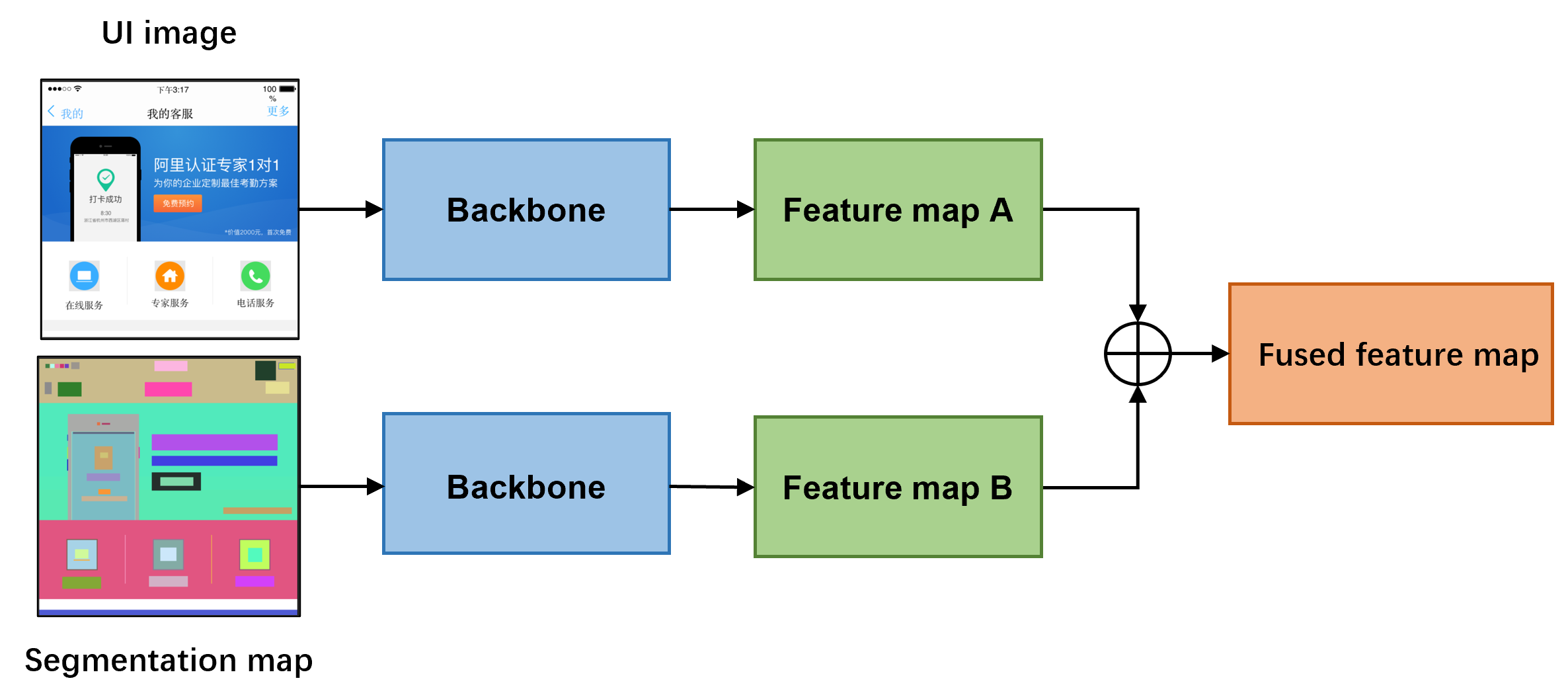}\\
{\footnotesize\sf (b) The feature fusion strategy.} \\
\end{tabular}
\caption{The two strategies for feature fusion.}
\label{fig:fusion}
\end{figure*}

There are two strategies for utilizing the spatial features based on the segmentation map. As shown in Fig.~\ref{fig:fusion}, we first generate the segmentation map which only contains boundary information. In the spatial fusion strategy, we stack the segmentation map to the original UI image to produce the fusion image. The fusion image contains the specific boundary of UI component. The fusion images are fed into the CNN backbone to produce feature map. In the feature fusion strategy, inspired by \citep{xu2017deep}, we use the high-dimensional features of spatial information as prior knowledge. Specifically, two high-dimensional feature maps are extracted from the original image and the corresponding segmentation map respectively by the CNN backbone, and then concatenated as a fused feature map. For hard examples, the fusion strategy can help the backbone to learn its boundary information more easily.

% fig fusion strategy (a)&(b)

\subsection{Dynamic Data Augmentation}

Training an effective object detection model for visual understanding requires massive high-quality data. Traditional methods of data augmentation for object detection are mainly image processing, including random flipping, random cropping, and resizing. These methods often cannot be dynamically adjusted to the data. Static data augmentation does not work well for small or large aspect ratio objects in the UI design scenario. Furthermore, training the MAD requires a large number of annotated UI design drafts. However, there is so far no such type of open dataset, while collecting such UI design drafts is quite time and effort consuming.

\begin{figure*}
\centering 
\includegraphics[scale=0.5]{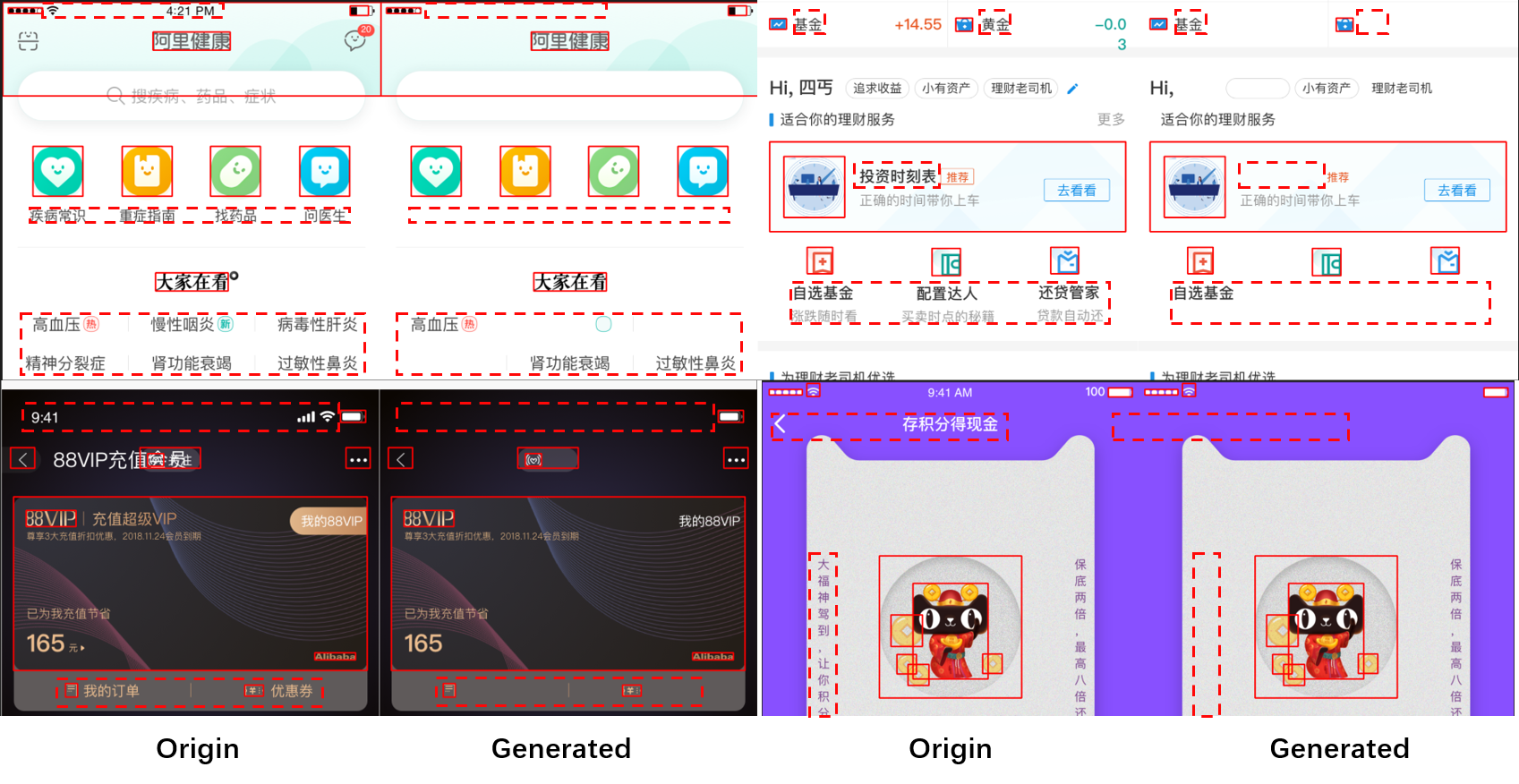}
\caption{Examples of data augmentation (solid box: preserved region, dashed box: removed region).} 
\label{fig_dataaug} 
\end{figure*}

In this study, we propose the dynamic data augmentation method to generate training data using existing Sketch files. Because the size of UI layers ranges widely, our data augmentation approach focuses on small components and large aspect ratio components. We present the dynamic data augmentation algorithm. Given the screenshot and associated JSON file, we traverse the layers list and randomly remove the layers that should not be merged according to a predefined ratio. So we generated a new JSON file and corresponding screenshot. Note that, the layers are randomly deleted at each training epoch. So we have a very diverse training sample. As shown in Fig.~\ref{fig_dataaug}, the meaning of "dynamic" is to keep the merging layers which are shown in the red solid-line box, and the other irrelevant layers shown in the red dashed box are randomly removed with a certain probability.

% fig6 data aug

\section{Experiments}

\subsection{Experimental Setting}

\noindent \textbf{Dataset.} The dataset contains screenshots of artboards and the corresponding JSON files with UI hierarchies. It is worth of mentioning that large-scale artboards may exceed the GPU memory limit. So we split each artboard into several images with a fixed size based on the shorter side. Because the shorter side of these artboards is not fixed, we resize the images such that the shorter side has a maximum of 800 pixels and the larger side has 1333 pixels. When dividing the dataset into the training set and test set, the images from the same artboard are divided into the same set to avoid introducing bias. Duplicated screenshots, which are produced from highly similar design drafts, are removed from the dataset. A total of 7399 screenshots are collected for experiments. In order to make the model focus on the hard examples, we take 30\% of the collected data for augmenting small objects which are defined as less than $32 \times 32$ pixels, and take 30\% of the original data for augmenting large aspect ratio objects when the aspect ratio is greater than three. In total, 5981 screenshots are utilized as the training set and augmented screenshots are added to the training set. The number of augmented screenshots generated by dynamic data augmentation is 5448. The test is performed on the remaining 1418 collected screenshots without data augmentation.

%Then images are resized such that the shorter side has a maximum of 800 pixels and the larger side has 1333 pixels. 

\noindent \textbf{Implementation details.} We use ResNet50 \citep{he2016deep} pre-trained on ImageNet as the backbone network. The FPN network \citep{lin2017feature} are used to extract a pyramid of features. In the first stage of the adaptive region proposal network, the anchor-free metric for sample discrimination with the thresholds of the center-region $\sigma_{ctr}$ and ignore-region $\sigma_{ign}$, is 0.2 and 0.5. In the second stage, we use the anchor-based metric with the IoU threshold of 0.7. The multi-task loss is set with the stage-wise weight $\alpha_1 = \alpha_2 = 1$ and the trade-off $\lambda$ = 10. The NMS threshold is set to 0.7. In the cascade box regression network, there are three stages with $IoU = {0.5, 0.6, 0.7}$. In the first stage, the input to the regressor is the adaptive region proposal from cascade region proposal network. At the following stages, re-sampling is implemented by simply using the regressed outputs from the previous stage. We follow the standard settings as in \citep{vu2019cascade} and \citep{vu2019cascade}. We implement our method with Pytorch and MMDetection codebase \citep{chen2019mmdetection}. We train all models with a mini-batch of 8 for 12 epochs using the SGD optimizer with a momentum update of 0.9 and a weight decay of 0.0001. The learning rate is initialized to 0.01 and divided by 10 after 8 and 11 epochs. It takes about 4 hours for the models to converge on an NVIDIA GeForce RTX 3090 GPU.

\noindent \textbf{Evaluation metrics.} We report performance with the metrics used in COCO detection evaluation criterion \citep{lin2014microsoft} and provide mean Average Precision (mAP) across various IoU thresholds i.e. IoU $= \left \{0.50 : 0.95, 0.5, 0.75\right \}$ and various scales: $\left \{small,medium,large \right \}$.

\subsection{Results}

\noindent \textbf{Detection performance comparison with baselines.} We first present the merging areas detection performance. Table~\ref{tab1} shows the performance comparison with the baselines. Note that our MAD uses the proposed spatial fusion strategy and dynamic data augmentation. It can be seen that our MAD is much better than the baselines. Specifically, the performance of our MAD is improved by an mAP of 14.5\% and 10.6\% compared to RetinaNet and Faster-RCNN respectively. It is effective to incorporate boundary prior knowledge to condition the bounding features. Our approach also outperforms competing methods such as Cascade-RCNN and GA-Faster-RCNN by an mAP of 7.8\% and 9.3\%. It shows that the feature alignment using adaptive anchors and the progressive refinement of boundaries can contribute to a performance boost. Compared to others methods, MAD enhances feature representation by fusing a prior knowledge of layers' boundary information. It also implies that MAD is especially good at detecting for UI layout.

% table1 
\begin{table*}[thp]\footnotesize
\centering
\caption{Performance Comparison with Baselines}
\label{tab1}
\addtolength{\tabcolsep}{4.8pt}
\begin{tabular*}{13cm}{lcccccc}
    \toprule[0.75pt]
    Method & $AP$ & $AP_{50}$ & $AP_{75}$ & $AP_S$ & $AP_M$ & $AP_L$\\
    \midrule[0.5pt]
    RetinaNet  & 0.545 & 0.708 & 0.602 & 0.523 & 0.634 & 0.395\\

    Faster-RCNN  & 0.584 & 0.726 & 0.647 & 0.588 & 0.627 & 0.430\\

    GA-Faster-RCNN  & 0.597 & 0.739 & 0.654 & 0.605 & 0.637 & 0.429\\

    Cascade-RCNN  & 0.612 & 0.734 & 0.665 & 0.612 & 0.657 & 0.456\\

    CRPN-Faster-RCNN  & 0.638 & 0.766 & 0.696 & 0.638 & 0.687 & 0.487\\

    MAD   & \textbf{0.690} & \textbf{0.801} & \textbf{0.753} & \textbf{0.677} & \textbf{0.752} & \textbf{0.536}\\
    \bottomrule[0.75pt]
\end{tabular*}
\end{table*}

\noindent \textbf{Comparison of two fusion strategies.} We conducted experiments to evaluate the effectiveness of the two fusion strategies separately. Table~\ref{tab2} shows that the spatial fusion strategy that fusing the information of layers' boundary into the original UI image is a better choice. A reasonable explanation is that the semantic information in the segmentation map is poor, thus it is difficult to enrich the features extracted from UI images. The fused features may even corrupt the original UI image feature representation, degrading model performance. The spatial fusion strategy enrich the original image with spatial features from layer boundaries at pixel level, which appears to be more efficient for the backbone to extract the semantic features. The results show that the spatial fusion strategy on all three models improves the mAP by at least 4.6\% over the feature fusion strategy.

% table2
\begin{table*}[thp]\footnotesize
\centering
\caption{The comparison of two fusion strategies}
\label{tab2}
\addtolength{\tabcolsep}{4.8pt}
\begin{tabular*}{13.5cm}{lcccccc}
    \toprule[0.75pt]
    Method & $AP$ & $AP_{50}$ & $AP_{75}$ & $AP_S$ & $AP_M$ & $AP_L$\\
    \midrule[0.5pt]
    RetinaNet+FF & 0.556 & 0.745 & 0.633 & 0.553 & 0.636 & 0.272\\
    RetinaNet+SF & \textbf{0.602} & \textbf{0.754} & \textbf{0.679} & \textbf{0.580} & \textbf{0.682} & \textbf{0.435}\\
    \midrule[0.5pt]
    \midrule[0.5pt]
    Faster-RCNN+FF & 0.495 & 0.690 & 0.583 & 0.505 & 0.563 & 0.170\\
    Faster-RCNN+SF & \textbf{0.622} & \textbf{0.757} & \textbf{0.697} & \textbf{0.614} & \textbf{0.683} & \textbf{0.442}\\
    \midrule[0.5pt]
    \midrule[0.5pt]
    MAD+FF  & 0.607 & 0.763 & 0.680 & 0.611 & 0.659 & 0.378\\
    MAD+SF  & \textbf{0.674} & \textbf{0.797} & \textbf{0.741} & \textbf{0.661} & \textbf{0.736} & \textbf{0.524}\\
    \bottomrule[0.75pt]
    \multicolumn{4}{p{8cm}}{\scriptsize FF denotes feature fusion and SF denotes spatial fusion}
\end{tabular*}
\end{table*}

\begin{figure}
\centering 
\includegraphics[scale=0.55]{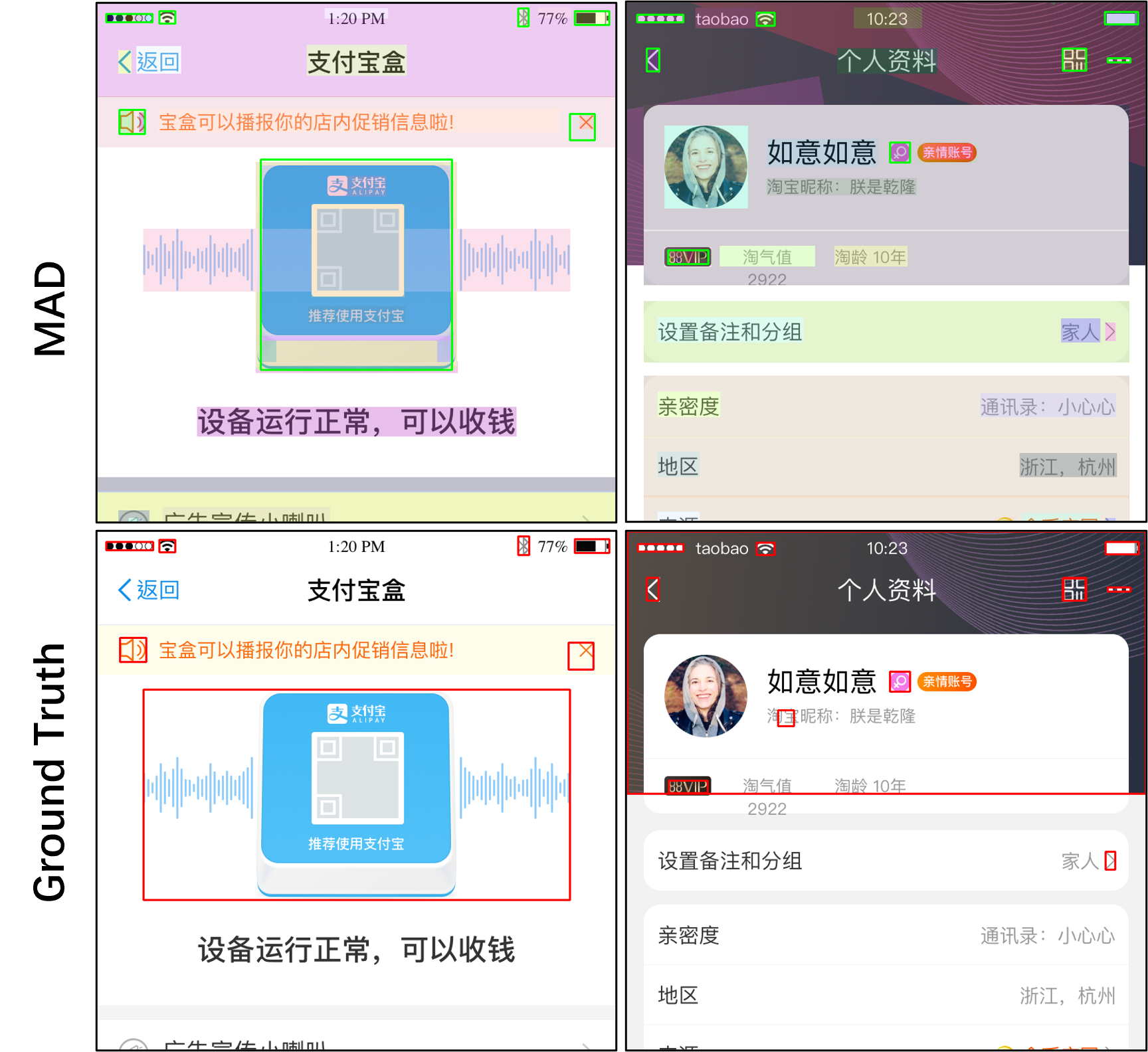}
\caption{Failure cases. Ground truth is shown in red boxes, while the predictions are shown in green boxes.}
\label{fig_failurecase} 
\end{figure}

\noindent \textbf{Failure cases analysis.} Our MAD performs not well for complex background components and some complicated shapes of UI components. Fig.~\ref{fig_failurecase} shows the typical examples that make the detection of merging layers very challenging. For UI components with complex shapes, MAD cannot accurately determine the boundaries of the component. For example, the predicted bounding box and the ground truth have very huge gap like the left side of Fig.~\ref{fig_failurecase}. So the merging algorithm cannot merge all fragmented layers into the right UI component. Another challenge is that the model has difficulty of learning salient visual features in complex design scenarios. For example, the designer design a background UI component like the right side of Fig.~\ref{fig_failurecase}. In this case, MAD cannot accurately determine the boundaries that belong to the background.
% fig failure case

\noindent \textbf{Layers Merging performance.} To evaluate the layers merging result, we define a metric called mean layers IoU which is similar to IoU. It is a score from 0 to 1 that specifies the intersection of prediction layer and the ground truth layer in the UI component. For example, the layers in a predicted group is 3 and the layers in ground truth is 4, then the layers IoU is 0.75. We average the sum of all layers IoU. And we perform operations on all labeled data to find the layers should be merged. Our model achieves 87.70\% in term of mean layers IoU. It implies that the layers merging algorithm can achieve a decent accuracy.

\subsection{Ablation Study}

\noindent \textbf{Data augmentation.} We investigate the contribution of proposed data augmentation approach. "MAD+DataAug" denotes the adoption of the data augmentation approach to MAD. "MAD+SF+DataAug" denotes the adoption of both the data augmentation approach and the spatial feature fusion strategy to MAD. From Table~\ref{tab3} we can observe that "MAD+SF+DataAug" improves the mAP by 1.6\% from "MAD+SF". It indicates the effectiveness of data augmentation for merging areas detection. Because the training set with the augmented data has more diverse samples for small and large aspect ratio objects, which increases the scalability of the model.

% table3
\begin{table*}[thp]\footnotesize
\centering
\caption{Ablation studies}
\label{tab3}
\addtolength{\tabcolsep}{4.8pt}
\begin{tabular*}{13cm}{lcccccc}
    \toprule[0.75pt]
    Method & $AP$ & $AP_{50}$ & $AP_{75}$ & $AP_S$ & $AP_M$ & $AP_L$\\
    \midrule[0.5pt]
    MAD & 0.651 & 0.778 & 0.705 & 0.658 & 0.688 & 0.512\\
    MAD+SF & 0.674 & 0.797 & 0.741 & 0.661 & 0.736 & 0.524\\
    MAD+DataAug & 0.659 & 0.775 & 0.706 & 0.666 & 0.704 & 0.519\\
    MAD+SF+DataAug & \textbf{0.690}  & \textbf{0.801} & \textbf{0.753} & \textbf{0.677} & \textbf{0.752} & \textbf{0.536}\\
    \bottomrule[0.75pt]
\end{tabular*}
\end{table*}

\noindent \textbf{Spatial fusion.} We conduct experiments with fusion strategy and without fusion strategy. It is worthy of mentioning that we used the spatial fusion strategy according to the previous experimental results. Table~\ref{tab3} indicates that after encoding the boundary information, MAD has 2.3\% mAP improvement. The obvious gain brought by the spatial fusion strategy suggests the necessity of the bounding spatial prior knowledge which enriches the feature representation. Besides, applying the fusion strategy to Faster-RCNN and RetinaNet all yielded 5.7\% and 3.8\% mAP improvements. This also demonstrates that the prior knowledge of boundary information can boost the performance of UI component detection models.

\begin{figure}
\centering 
\includegraphics[scale=0.45]{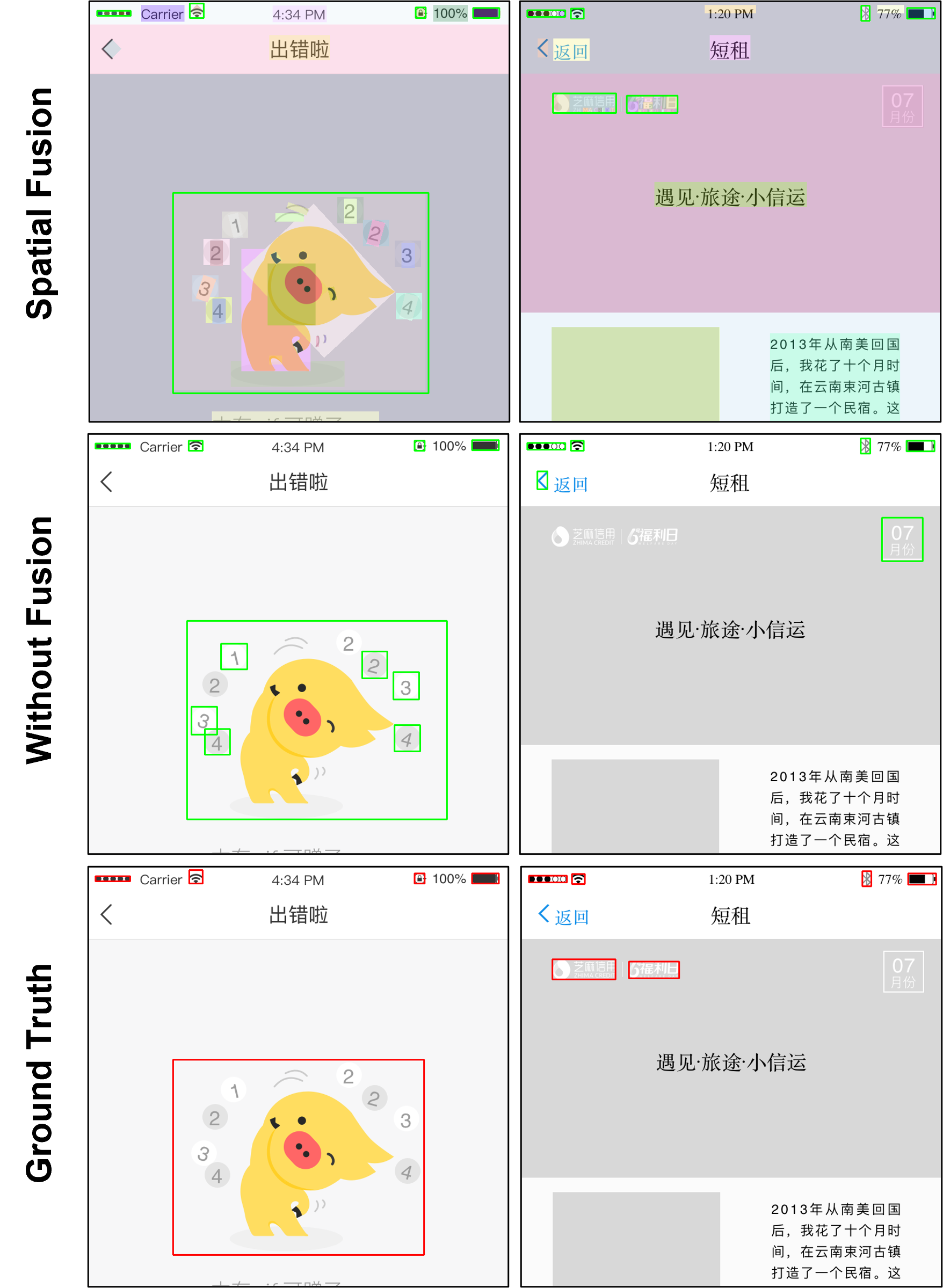}
\caption{Examples of spatial fusion. Ground truth is shown in red boxes. The predictions are shown in green boxes.} 
\label{fig_fusioncCase} 
\end{figure}

Fig.~\ref{fig_fusioncCase} presents the advantage of spatial fusion with examples. As shown in the right side of Fig.~\ref{fig_fusioncCase}, it shows that the spatial fusion strategy adds a clear boundary to distinguish different UI components areas. Thus the model can visually identify the region as a combination of multiple layers more easily and increase the prediction performance of layers merging. Furthermore, in a real design scenario, the designer may have already merged some layers like the left side of Fig.~\ref{fig_fusioncCase}. So there is no need to detect the digital icons. But there is no information here to prompt the model. This could cause the model to wrongly recognize the region of UI component. The result shows that MAD with spatial fusion strategy detects the UI component successfully. Because the model can learn visual boundary features to avoid detecting UI components that have been merged.

% fig fusion case

\section{User Study and Application}

The goal of this work is to automatically merge fragmented layers in UI design drafts to reduce the time developers spend on understanding and modifying code, and improve development efficiency. The extensive experiments above demonstrate our model outperforms other baselines with decent advance. However, the satisfaction of the generated code might be subjective depending on different users or developers. In this section, we build a complete pipeline for merging fragmented layers called UILM. Specifically, a screenshot with view hierarchy is fed into parsing pipeline to produce a segmentation map. Then the screenshot and corresponding segmentation map are fused together. We feed it into our MAD to predict the bounding boxes as detected merging areas. Our layers merging algorithm takes the merging areas and searches out the associated layers to be merged. To better evaluate the usefulness of UILM, we conduct a user study to investigate the feedback from developers. We also apply UILM in Taobao's front-end development process.

\subsection{Evaluation Metrics}

There are no existing evaluation metrics for code generation from design drafts in the literature. Inspired by the GUI design evaluation\citep{zhao2021guigan}, we propose two novel objective metrics for participants to estimate the quality of code generation by considering the characteristics of front-end code implementation: code availability and code modification time. Besides, to confirm if UILM improves code readability and maintainability, we propose two subjective metrics by asking participants to rate their experience. The four metrics are quantified by scores ranging from one to five. Code availability is used to evaluate how much generated code is available for production. We use git service\footnote{\url{https://git-scm.com/}} to record the lines of code changes. The calculation approach is as follows:
\begin{equation}
    availability= 1-\frac{lines\ of\ code\ changes}{total\ lines\ of\ code}
\end{equation}
% table3

\begin{table}[htbp]\footnotesize
\centering
\caption{Evaluation metric for code generation}
\label{tab:metric}
\begin{tabular}{ccc}
    \toprule[0.75pt]
    Score & Code availability(a) & Code modification time(t) \\
    \midrule[0.5pt]
    1 & $a<0.75$ & $t \ge 10min$\\
    2 & $0.75 \leq a < 0.80$ & $8min \leq t < 10min$\\
    3 & $0.80 \leq a < 0.85$ & $6min \leq t < 8min$\\
    4 & $0.85 \leq a < 0.90$ & $4min \leq t < 6min$\\
    5 & $a \ge 0.90 $ & $t < 4min$\\
    \bottomrule[0.75pt]
\end{tabular}
\end{table}
For data statistics, the metric score is defined as shown in Table~\ref{tab:metric}. A score of one to five corresponds to code availability between 0.75 and 1.0. Code modification time is to evaluate the time required to adjust the code to actual production standards. The participants were recorded the time of modifying the generated code. We use two minutes as an interval. The modification time more than ten minutes as one score, and within four minutes as five scores. The scores of readability and maintainability are given by participants from one to five subjectively, representing the quality of code from low to high.

\subsection{Procedures}

In this study, it is defined three categories of common UI components that contain fragmented layers, which are icon, atmosphere UI, background UI. We fetch the corresponding components from design drafts and then generate the code using Imgcook. For each category, we have ten samples. It is worth noting that we do not remove the components overlapped on the background UI to ensure generalization.

We recruited frond-end engineers for code evaluation, who have more than three years programming experience and at least two years front-end development experience by using Vue\footnote{\url{https://vuejs.org/}} which is a progressive JavaScript framework. They are introduced to a detailed explanation about the code evaluation metrics. Then they were provided with generated code. We calculated the scores of modification time and amount of code changes, and collected their rated scores of code readability and maintainability. Note that they were not aware of which code is merged by our method and all of them evaluated the generated code without any discussion. After the experiment, we asked the participants to leave some feedback about our UILM.

\subsection{Results}

As shown in Table~\ref{tab:HumanEvaluation}, the code generated after merging the fragmented layers using our UILM outperforms the code without merging in average code availability, modification time, readability and maintainability. In addition to the average score, our method also outperforms the method without merging, on four metrics for all three UI component categories. The result demonstrates the generalization of our UILM. We analyze the experimental results in detail. The merged icon and atmosphere UI significantly decrease the code modification time, since our UILM is good at detecting icon and atmosphere UI. Besides, the code readability and maintainability of background UI are significantly improved by 70.27\% and 48.28\% respectively. Regarding the background UI, UILM not only merged the fragmented layers of the detected background, but also merge the overlapping icons and atmosphere UI. Hence, the improvement is obvious and significant. It is proven that our method can help improve the quality of the generated code.

To demonstrate the significance of UILM, we carry out the Mann-Whitney U test \citep{fay2010wilcoxon} which is specifically designed for small samples. $p<0.05$ is typically considered to be statistically significant and $p<0.01$ is considered to be highly statistically significant. The result shows that our UILM can contribute significantly to the code generation in all four metrics except the code availability in the background UI. Besides, the participants concluded that merging fragmented layers has a positive effect on improving generated code quality. They thought generated code without merging tended to have redundant containers, but most of containers can be fixed after merging. One of the participants emphasized that UILM had the potential to aid the automatic code generation process: "The key requirement is fast iteration. A developer could generate the code with clean layout is very beneficial to maintain and modification." 
\begin{table*}[thp]
  \centering  
%   \fontsize{8}{7}\selectfont
  \addtolength{\tabcolsep}{0.5pt}
  \begin{threeparttable}  
  \caption{Performance of Human Evaluation} \label{tab:HumanEvaluation}
    \begin{tabular}{lllll} 
    \toprule
    \multirow{2}{1in}{Category}&  \multirow{2}{0.6in}{Metric}&\multicolumn{3}{c}{Score}\cr
    \cmidrule(lr){3-5}
    &&Non-merge&Merge&Increment\cr
    \midrule
    \multirow{4}{1in}{Icon}
    & code availability &$2.72$&$3.54^{**}$&+30.15\% \\
    & code modification time &$2.80$&$3.78^{*}$&+35.0\% \\
    & readability &$3.84$ &$4.44^{*}$&+15.63\% \\
    & maintainability &2.98&$4.06^{*}$&+26.60\%\cr
    \cmidrule(lr){1-5}
    \multirow{4}{1in}{Atmosphere} 
    & code availability & $2.72$ & $3.50^{*}$&+28.68\% \\
    & code modification time & $2.62$ & $3.68^{**}$&+40.46\% \\
    & readability &  $3.14$ & $4.20^{*}$&+33.76\% \\
    & maintainability&2.82&$3.82^{*}$&+35.46\%\cr
    \cmidrule(lr){1-5}
    \multirow{4}{1in}{Background} 
    & code availability &1.94 &2.46 &+26.80\% \\
    & code modification time &2.24 &$2.98^{*}$&+33.04\% \\
    & readability &$2.22$ &$3.78^{**}$&+70.27\% \\
    & maintainability&2.32&$3.44^{*}$&+48.28\%\cr
    \cmidrule(lr){1-5}
    \multirow{4}{1in}{\textbf{Average}}
    & code availability &$2.46$ & 3.17&+28.86\%\\
    & code modification time  &$2.55$ &3.48&+36.47\% \\
    & readability &$3.07$ &4.14&+34.85\% \\
    & maintainability&$2.71$ &3.77&+50.48\%\cr
    \bottomrule{\scriptsize ** denotes $p < 0.01$ and * denotes $p < 0.05$}
    \end{tabular}
    \end{threeparttable}
\end{table*} 

\subsection{Application}

\begin{figure}
\centering 
\includegraphics[scale=0.46]{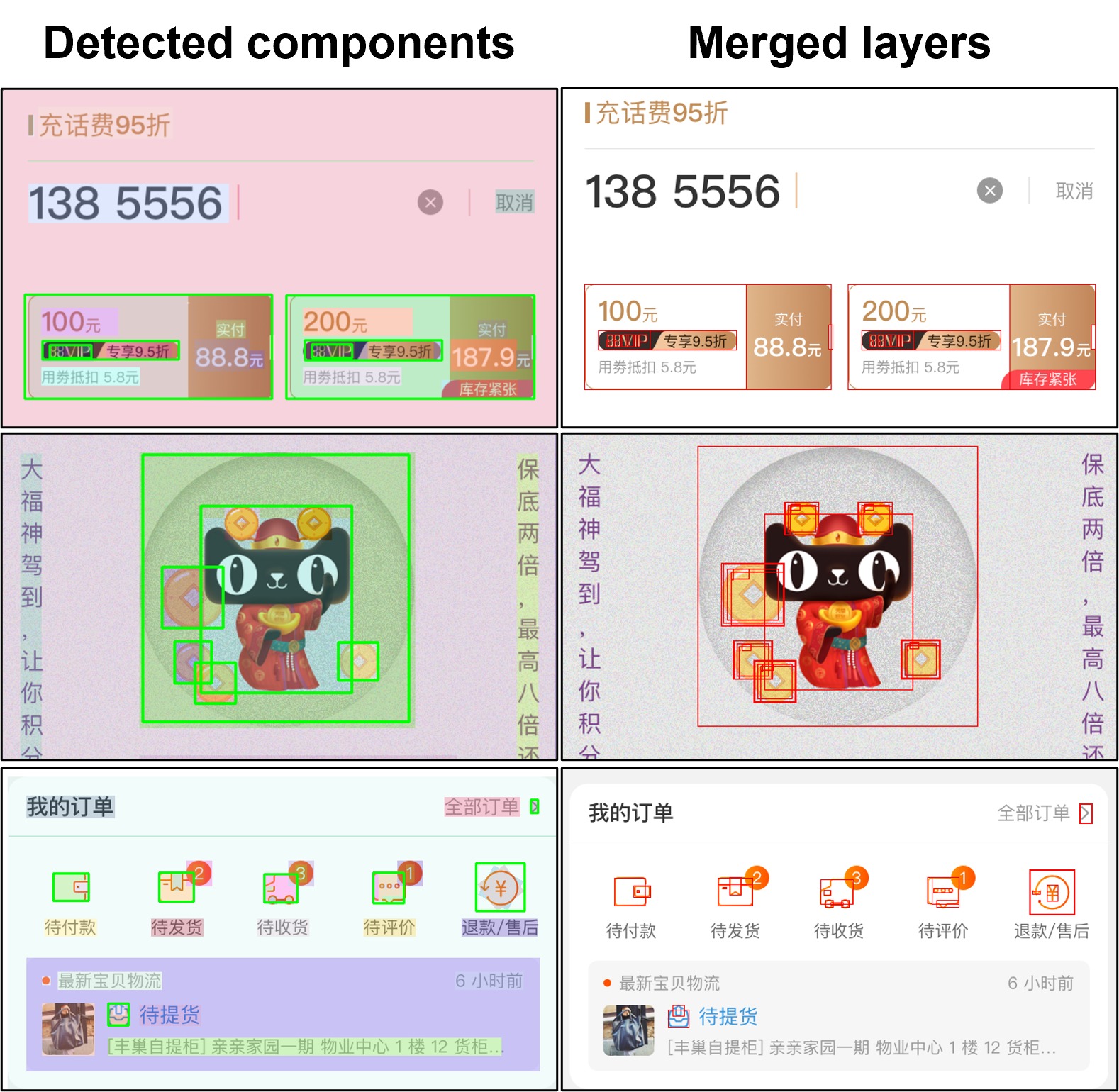}
\caption{Examples of associative layer merging results. The green box represents the detected components. The red box represents the layers to be merged} 
\label{fig_applicationcase} 
\end{figure}

\begin{figure}
\centering 
\includegraphics[scale=0.55]{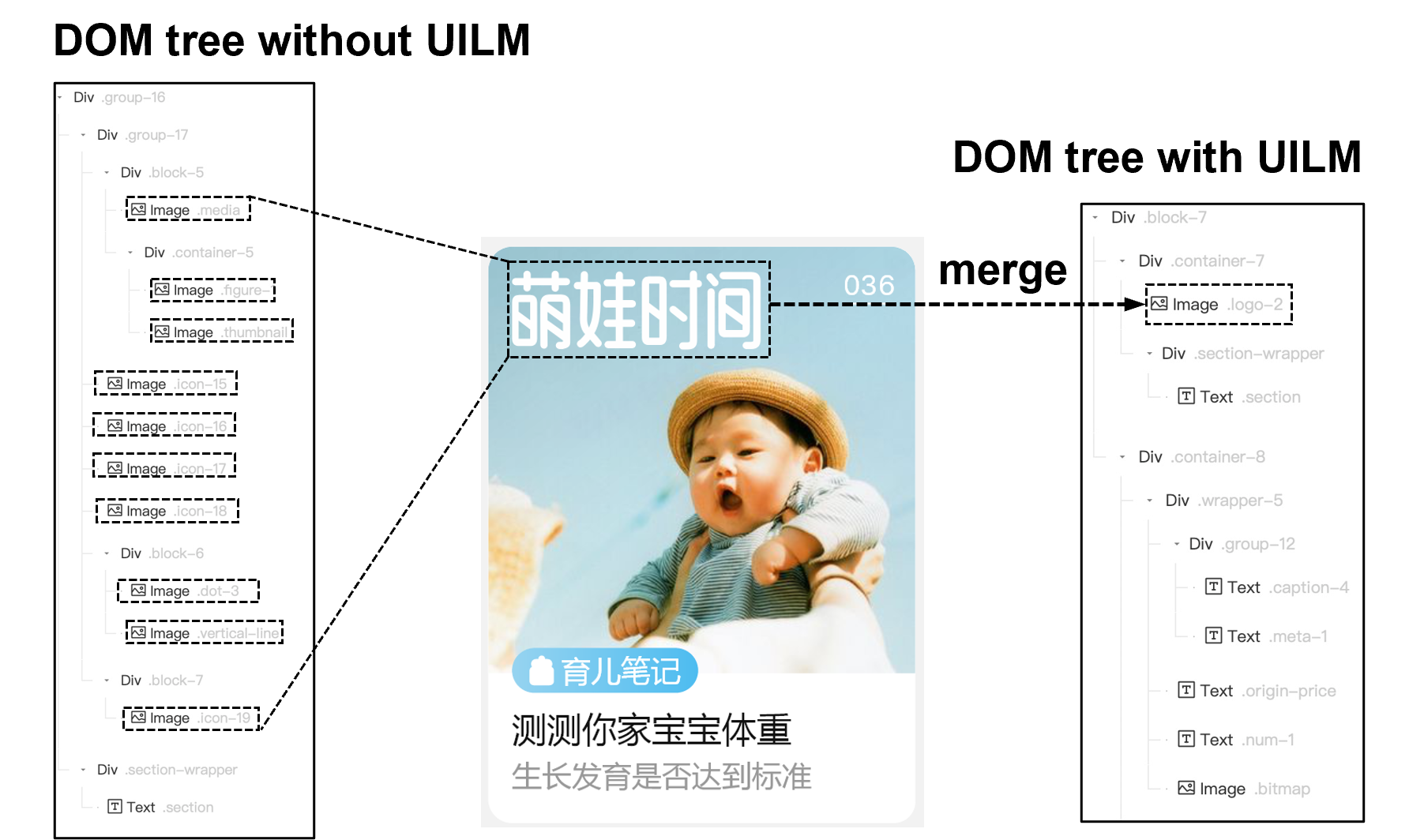}
\caption{An example of generated DOM tree with and without UILM. In the DOM tree without UILM, the dashed boxes represent redundant containers. In the DOM tree with UILM, the single image container represents the "text-like" UI component} 
\label{fig_applicationcode} 
\end{figure}

To understand the applicability of UILM from an industrial prospective, we apply our method to Taobao's front-end development process. We first invited Taobao's front-end engineers working on automated code generation development to provide some typical and challenging UI design drafts to be merged in practical application scenarios. To evaluate these representative samples, our model as a plugin in a code generation tool automatically merges fragmented layers in UI design drafts. We visualize some challenging detection results as show in Fig.~\ref{fig_applicationcase}. The green box represents the detected components by MAD in the first column and the red box represents the layers to be merged in the second column. The results show that the proposed method detects all components successfully and merges all associated layers. It also demonstrates our UILM can process various UI components such as various icons and nested components. We also display a DOM tree associated with generated front-end code. Fig.~\ref{fig_applicationcode} shows that without merging fragmented layers, the "text-like" logo consists of ten image containers in the generated DOM tree. Its nested structure and the redundant containers result in the poor readability and maintainability of generated code. Our UILM can merge these layers into one group with a "\#merge\#" tag for the recognition by downstream code generation algorithms. When recognizing the annotation, the automatic code generation tool, such as Imgcook, can merge all layers inside the group to produce a singe image container. The structure of generated DOM tree with our UILM is simplified and significantly improve the quality of generated code with a cleaner DOM tree.

\section{Conclusion}

This paper investigates a novel issue about layers merging in an automatic design draft to UI view code process, which can decrease the quality of generated code. To solve this issue, we innovatively proposed UILM by detecting the areas of UI components and merging the fragmented layers to UI components. By incorporating boundary prior knowledge, the MAD can achieve more than 6.8\% boost in detection mAP compared with the best baseline. We also propose a dynamic data augmentation approach to boost the performance of MAD. As the first work of merging fragmented layers in UI design drafts, we construct a large well-annotated UI dataset to train our model and evaluate the effectiveness of our method. Furthermore, UILM is also proven to be effective in real practice by building a test pipeline.

\bibliographystyle{unsrtnat}
\bibliography{references}  %%% Uncomment this line and comment out the ``thebibliography'' section below to use the external .bib file (using bibtex) .





% \end{thebibliography}

\end{document}